\documentclass[10pt,twocolumn,letterpaper]{article}

\usepackage{cvpr}
\usepackage{times}
\usepackage{epsfig}

\usepackage{graphicx}
\usepackage{subcaption}
\usepackage{amsmath}
\usepackage{amssymb}
\usepackage{booktabs}
\usepackage{multicol}

\usepackage[numbers,sort,compress]{natbib}

\usepackage[pagebackref=true,breaklinks=true,letterpaper=true,colorlinks,bookmarks=false]{hyperref}
\usepackage{amsthm}
\theoremstyle{definition}

\usepackage{algorithmic}
\usepackage{algorithm}

\newlength\myindent 
\cvprfinalcopy 

\begin{document}

\title{Asynchronous Hierarchical Federated Learning}

\author{
Xing Wang\\
Splunk Inc.\\
{\tt\small xingw@splunk.com}
\and 
Yijun Wang\\
University of Illinois\\
{\tt\small yijunw4@illinois.edu}


}

\maketitle

\begin{abstract}


Federated Learning is a rapidly growing area of research and with various benefits and industry applications. Typical federated patterns have some intrinsic issues such as heavy server traffic, long periods of convergence, and unreliable accuracy. In this paper, we address these issues by proposing asynchronous hierarchical federated learning, in which the central server uses either the network topology or some clustering algorithm to assign clusters for workers (i.e., client devices). In each cluster, a special aggregator device is selected to enable hierarchical learning, leads to efficient communication between server and workers, so that the burden of the server can be significantly reduced. In addition, asynchronous federated learning schema is used to tolerate heterogeneity of the system and achieve fast convergence, i.e., the server aggregates the gradients from the workers weighted by a staleness parameter to update the global model, and regularized stochastic gradient descent is performed in workers, so that the instability of asynchronous learning can be alleviated. We evaluate the proposed algorithm on CIFAR-10 image classification task, the experimental results demonstrate the effectiveness of asynchronous hierarchical federated learning.
\end{abstract}

\section{Introduction}


Federated learning is a machine learning methodology where the model updates happen in many clients (mobile devices, silos, etc.), as opposed to a more traditional setting, where training happens on a centralized server. This approach has many benefits, such as avoiding the upkeep of a centralized server, minimizing the network traffic between many clients and servers, as well as keeping sensitive user data anonymous and private. 
For cross-device Federated Learning, the most common architecture pattern is a centralized topology \cite{2016arXiv160205629B}. This topology has been a popular area of research, and has seen much success when applied to commercial products. However, the centralized topology introduces some challenges.

We are learning from the tradeoffs presented by the comparison of certain topologies and help us focus on which tradeoffs we want to measure specifically.
In a centralized network topology, a central server is where all the training happens. However, in some scenarios, a central server may not always be desirable; or the central may not be powerful enough \cite{2016arXiv161005202V}. The central server could also become a bottleneck for large amount of network traffic, and large number of connections. This bottleneck is exacerbated when the federated networks are composed of a massive number of devices \cite{2017arXiv170509056L}. Furthermore, current Federated Learning algorithms, such as Federated Averaging, can only efficiently utilize hundreds of devices in each training round, but many more are available \cite{2019arXiv190201046B}.

One approach that attempted to address these challenges is decentralized training \cite{2016arXiv160205629B,2017arXiv170509056L,2018arXiv180307068T,2016arXiv161005202V}. However, in large scale FL systems, a fully decentralized FL topology is inefficient, since the convergence time could be long, and the traffic between devices could be too intensive. Well-connected or denser networks encourage faster consensus and give better theoretical convergence rates, which depend on the spectral gap of the network graph. However, when data is IID, sparser topologies do not necessarily hurt the convergence in practice. Denser networks typically incur communication delays which increase with the node degrees \cite{2019arXiv191204977K}. 

In this paper, we propose asynchronous hierarchical federated learning, in which the central server uses either the network topology or some clustering algorithm to assign clusters for workers (i.e., client devices). In each cluster, a special aggregator device is selected to enable hierarchical learning, leads to efficient communication between server and workers, so that the burden of the server can be significantly reduced. In addition, asynchronous federated learning schema is used to tolerate heterogeneity of the system and achieve fast convergence, i.e., the server aggregates the gradients from the workers weighted by a staleness parameter to update the global model, and regularized stochastic gradient descent is performed in workers, so that the instability of asynchronous learning can be alleviated.

The rest of this paper is organized as follows. Section \ref{sec:liter} discusses several recently developed federated learning work that are related to ours.  In Section \ref{sec:approach}, we illustrate our proposed algorithm in details, theoretical analysis is shown as well. We conduct experiments and show the evaluation results in Section \ref{sec:res}. Finally, we conclude our findings and discuss future work directions in Section \ref{sec:conc}.

\section{Related Work} \label{sec:liter}


The most widely used and straightforward algorithm to aggregate the local models is to take the average, proposed in \cite{mcmahan2017communication} and known as Federated Averaging (\textit{FedAvg}).
In FL, the communication cost often dominates the computation cost \cite{mcmahan2017communication}, thus is one of the key issues we need to resolve for implementing FL system at scale. In particular, the state-of-the-art deep learning models are designed to achieve higher prediction performance at the cost of increasing model complexity with millions or even billions of parameters \cite{devlin2018bert, brown2020language}. On the other hand, FL requires frequent communication of the models between the server and workers. As such, \textit{FedAvg} \cite{mcmahan2017communication} encourages each worker to perform more iterations of local updates before communicating during global aggregation, this results in significantly less communication rounds, and also increases the accuracy eventually as model averaging produces regularization effect. Another way to decrease the communication cost is to reduce the size of model information that needs to be sent, either through model compression techniques such as sparsification \cite{stich2018sparsified} and quantization \cite{caldas2018expanding}, or only select a small portion of important gradients to be communicated \cite{tao2018esgd} based on the observation that most of deep learning model parameters are closed to zero \cite{strom2015scalable}.  However, these methods may result in deterioration of model accuracy, or incur high computation cost \cite{lim2020federated}. Alternatively, \cite{liu2020client} proposed client-edge-cloud hierarchical federated learning (\textit{HierFAVG}), an edge computing paradigm in which the edge servers play the roles of intermediate parameter aggregators. The hierarchical FL algorithm leverages on the proximity of edge servers, significantly relieves the burden of the central server on remote cloud. 
\cite{yuan2020hierarchical} introduces a hierarchical federated learning protocol through LAN-WAN orchestration, which involves a hierarchical aggregation mechanism in the local-area network (LAN) due to its abundant bandwidth and almost negligible monetary cost than WAN, and incorporates cloud-device aggregation architecture, intra-LAN peer-to-peer (p2p) topology generation, inter-LAN bandwidth capacity heterogeneity.
While the hierarchical learning pattern is promising to reduce communication, it is not applicable to all networks, as the physical hierarchy may not exist or be known \textit{a priori} \cite{li2020federated}. 
\cite{sattler2020clustered} designs and implements Clustered Federated Learning (CFL) using a cosine-similarity-based clustering method that creates a bi-partitioning to group client devices with the same data generating distribution into the same cluster.  Client devices are clustered into different groups according to their properties. It has better performance for the non-IID-severe client network, without accessing the local data.
\cite{briggs2020federated} implemented a hierarchical clustering step (FL+HC) to separate clusters of clients by the similarity of their local updates to the global joint model. Once separated, the clusters are trained independently and in parallel on specialised models.
In \cite{nguyen2020self}, a self-organizing hierarchical structured FL mechanism is implemented based on democratized learning, agglomerative clustering, and hierarchical generalization.

Most of the current FL systems are implemented using synchronous update, which is susceptible to the straggler effect. To address this problem, \cite{xie2019asynchronous} proposed an asynchronous algorithm for federated optimization, \textit{FedAsync}, which solves regularized local problems and then uses a weighted average with respect to the staleness to update the global model. \cite{chen2019asynchronous} presented \textit{ASO-Fed} for asynchronous online FL, which uses the same surrogate objective for local updates, while the local learning rate is adaptive to the average time cost of past iterations. This type of surrogate of adding such a proximal term for local updates was introduced in \textit{FedProx} \cite{li2020fedhetero}, which mainly aimed to address the problem of system heterogeneity, yet it turns out the similar idea can be adopted well to asynchronous federated learning.


There also exist several work that focused on decentralized solutions. Gossip learning is a decentralized alternative to federated learning.   \cite{hu2019decentralized} adopted gossip learning without aggregation servers nor a central component. Knowing that peer-to-peer bandwidth is much smaller than the worker’s maximum network capacity, the system could fully utilize the bandwidth by saturating the network with segmented gossip aggregation and the experiments showed that the training time can be reduced significantly with great convergence performance. 
\cite{hegedHus2019gossip} presents a thorough comparison of Centralized Federated Learning and Gossip Learning. Examine the aggregated cost of machine learning in both cases, considering also a compression technique applicable in both approaches.
\cite{lalitha2019peer} presents a peer-to-peer Federated Learning on graphs, which is a distributed learning algorithm in which nodes update their belief by judicially aggregating information from their local observational data with the model of their one-hop neighbors to collectively learn a model that best fits the observations over the entire network.
Coral is a peer-to-peer self-organizing content distribution system introduced by \cite{freedman2003sloppy}. Coral creates self-organizing clusters of nodes that fetch information from each other to avoid communicating with more distant or heavily-loaded servers. 
As a peer-to-peer FL framework particularly targeted towards medical applications, BrainTorrent introduced in \cite{roy2019braintorrent} presents a highly dynamic peer-to-peer environment, where all centers directly interact with each other without depending on a central body.

\section{Approach} \label{sec:approach}


\subsection{Problem Formulation}
We consider the supervised federated learning problem which involves learning a single global statistical model owned by the central server, while each of $N$ devices owns a private dataset and works on training a local model. Let $\mathbf{w}$ parameterize the model, and $\mathcal{D}^i = \{\mathbf{x}_j, y_j\}$ denote the training dataset owned by $i$-th device, where $i\in\{1, \cdots, N\}$, $\mathbf{x}_j$ is the $j$-th input sample from $\mathcal{D}^i$, while $y_j$ is the corresponding label. Denote $\ell (\mathbf{x}_j, y_j | \mathbf{w})$ as the loss function presents the prediction error, our overall goal is to minimize the empirical loss $\mathcal{L}(\mathbf{w})$ over all distributed training data $\mathcal{D} = \bigcup_{i=1}^n \mathcal{D}^i$, i.e., we aim at solving the following optimization problem:
\begin{equation}
\min\limits_\mathbf{w}  \mathcal{L}(\mathbf{w}) = \frac{ \sum_{i=1}^n \sum_{j\in \mathcal{D}^i} \ell (\mathbf{x}_j, y_j | \mathbf{w}) }{|\mathcal{D}|}.
\end{equation}
The problem is often solved by mini-batch stochastic gradient descent (SGD), in which during each step, the model is updated as
\begin{equation}
\mathbf{w} \leftarrow \mathbf{w} - \alpha \dfrac{\partial \mathcal{L}}{\partial \mathbf{w}} ,
\end{equation}
where $\alpha$ denotes the learning rate, and the average gradient 
\begin{equation}
\dfrac{\partial \mathcal{L}}{\partial \mathbf{w}}= \dfrac{1}{m} \sum_{j \in B} \dfrac{\partial \ell_j }{\partial \mathbf{w}} 
\end{equation}
is derived through back-propagation from the mini-batch $B$ of $m$ input samples. In the typical FL setting, each device $i$ performs SGD with data sampled from its own private training dataset $\mathcal{D}^i$ and train a local model 
\begin{equation}
\mathbf{w}_i = \text{arg}\min\limits_{\mathbf{w}_i} \mathcal{L}_i(\mathbf{w}_i) = \frac{ \sum_{j\in \mathcal{D}^i} \ell (\mathbf{x}_j, y_j | \mathbf{w}_i) }{|\mathcal{D}^i|} ,
\end{equation}
the server aggregates all local models collected from the workers and update the global model which is then sent back to the workers for next iteration. 

\subsection{Proposed Method} \label{sec:our_method}


\subsubsection{Initialization at Central Server}
We consider a hierarchcal FL system which has one central server on the cloud. The central server owns the global model $\mathbf{w}$ and denotes the timestamp $t$ for the model parameters. 
Therefore, as the learning begins, the central server initializes the global model parameters $\mathbf{w}$, its timestamp $t=0$, as well as several hyperparameters that are required by learning. 
In addition, given the network information of all client devices, we allow the central server to be responsible for the knowledge of the hierarchical communication topology. 
In the case of mobile edge computing, the partition and hierarchy can be naturally formed by the communication edges, as the links between the central server with the edge servers or base stations form a star topology, and so do the links between each edge server with the devices. We extend this architecture to a more general case by allowing the central server to run clustering algorithm to assign which cluster each device belongs to, as well as a special device in each cluster, which we denote as the ``\textit{aggregator}'', that plays the role of an edge server, that is, provides inter-hierarchy communication including downlink transmission from the central server and client devices and uplink transmission from the clients to the central server, also aggregates information to reduce the necessary communication. In FL, this aggregation work is specific to aggregating of the clients' updated weights/gradients, and sending them to the central server. The downlink transmission is straightforward: the server periodically sends the global model with timestamp as well as the hyperparameters for the learning task to the aggregators, and the aggregator serves as the parent node of the client devices in each cluster, forwards the information it receives from the central server to its children nodes.

\subsubsection{Learning on Local Clients}
Upon receiving the global model parameters $\mathbf{w}$ with its timestamp (according to the central server clock) from the central server, the worker client performs local update.
In order to mitigate the deviations of the local models on an arbitrary device $j$ from that of the central server, following \textit{FedAsync} \cite{xie2019asynchronous}, instead of minimization of the original local loss function $\ell_j$, client $j$ locally solves a regularized optimization problem, i.e., performs SGD update for one or multiple iterations on the following surrogate objective:
\begin{equation} \label{eqn:local}
 \min\limits_{\mathbf{w}_j} g_j(\mathbf{w}_j) =  \mathbb{E}_{(\mathbf{x}, y) \sim \mathcal{D}^j} \Big[\ell_j(\mathbf{w}_j) + \frac{\lambda}{2} || \mathbf{w}_j - \mathbf{w} ||^2 \Big],     
\end{equation}
in which the regularization term $\frac{\lambda}{2} || \mathbf{w}_j - \mathbf{w} ||^2$ controls the deviation of the local models. After local learning, the client sends its updated parameters as well as the original model timestamp to its parent node, the corresponding aggregator. The next local learning iteration will based on the newest received global model and the corresponding timestamp.

\subsubsection{Learning on Cluster Aggregators}
On an aggregator, the rate of receiving updates from the clients may vary caused by several reasons, such as heterogeneity of the computation power among devices, network delay, etc. We propose to perform asynchronous federated learning, that is, the aggregator immediately aggregates the update from the clients and reports to the central server. In real implementation, we can use a thread-safe FIFO queue to store the updates from the clients inside each aggregator, and periodically aggregates the results in the queue without waiting for that from some potential stragglers. This is different from the synchronous FL paradigm, and the uplink communication is then non-blocking. Again, following \textit{FedAsync} \cite{xie2019asynchronous}, we use a function of staleness to mitigate the error caused by obsolete models. Intuitively, more staleness results in larger error. On an aggregator device, assume the latest global model it received was with timestamp $t'$ (according to central server clock) at the moment it is about to aggregate the updates,
and the local model from client was with timestamp $t$, then it must be true that $t' \ge t$. We modify the learning rate to be weighted by the staleness:
\begin{equation}
\alpha_{t'} = \alpha \times \sigma(t'-t), 
\end{equation}
in which $\sigma(z)$ is the staleness function. Different forms of $\sigma(z)$ were defined in \cite{xie2019asynchronous}, such as:
\begin{itemize}
    \item the polynomial form: \begin{equation}  \label{eqn:poly_stale}
        \sigma (t'-t) = (t'-t+1)^{-\beta} ,
    \end{equation}
    \item the hinge form: 
    \begin{equation} 
    \sigma (t'-t) = \begin{cases}
    1      & \quad \text{if } t'-t \le b \\
    \dfrac{1}{a(t'-t-b)+1}  & \quad \text{otherwise}
  \end{cases} \end{equation}
\end{itemize}
Note that $\sigma(t'-t)=1$ if $t'=t$, and monotonically decreases as $t'$ and $t$ deviates more, so that the obsolete update would affect the model less as it shrinks the learning rate. 
Therefore, upon receiving local update from client device $j$, the model can then be updated on the cluster aggregator $k$ as:
\begin{equation}
 \mathbf{w}_k^{(t')} \leftarrow (1-\alpha_{t'}) \mathbf{w}^{(t')} + \alpha_{t'} \mathbf{w}^{(t)}_{j}.
 \end{equation}

Or equivalently, if we aggregate the gradients collected from the clients, we have:
\begin{equation} \mathbf{dw}_k^{(t')} = \sum \alpha \sigma(t'-t) \mathbf{dw}_j^{(t)},  \end{equation}
where $\mathbf{dw}_j^{(t)} $ is the gradient collected by device $j$ in $k$-th cluster. 

\subsubsection{Learning on Central Server}
The central server aggregates the results from the cluster aggregators to update the global model. Similar to the learning procedure on the cluster aggregators, in the central server, we can use a queue to store the updates from the aggregators. As asynchronous learning, the numbers of updates gathered from each of the aggregators can be imbalanced. As such, we let the aggregator $k$ send the number of updates $n_k$ along with the aggregated results (i.e., $\mathbf{w}_k^{(t')}$ and timestamp $t'$) to the central server. We assume the newest global model was updated at timestamp $t''$ according to the central server clock, and that it must hold that $t'' > t'$. Combine the update counts information and the staleness schema, by collecting the update from aggregator $k$, the learning rate is weighted and modified as $$\alpha_{t''} = \dfrac{n_k}{N} \sigma(t''-t') \alpha, $$
where $N$ is the total number of devices in the FL system which is known to the central server. 
Note that this weighting mechanism makes sense if the data are i.i.d. over the clients. However, the bias could be severe if non-i.i.d. data are involved, as the mechanism favors learning for faster computed and communicated devices, in which case we need to carefully tune and design a more complicated weighting mechanism. The central server updates the global model as follows:
\begin{equation}
    \mathbf{w}^{(t''+1)}  \leftarrow (1-\alpha_{t''}) \mathbf{w}^{(t'')} + \alpha_{t''} \mathbf{w}^{(t')}_{k},
 \end{equation}

The detailed algorithm is illustrated in Algorithm \ref{alg:fedah}.

\begin{algorithm}[htb!]
\caption{Asynchronous Hierarchical Federated Learning (FedAH)}
\label{alg:fedah}
\begin{algorithmic}[1]
\footnotesize
\STATE \underline{\textbf{Central Server:}}
    \STATE Assign clusters and aggregators according to network topology or by running clustering algorithm.
    \STATE Initialize global model $\mathbf{w}$ and time clock $t$.
\FOR{$t=0, \cdots, T-1$ \textbf{until} end of learning}
    \STATE Broadcast $(\mathbf{w}, t)$ to its direct children aggregators. 
    \STATE Receive triples $\big(\mathbf{dw}_k^{(t')}, t', n_k\big)$ from any direct child aggregator $k$.
    \STATE Update global model 
    $\mathbf{w}^{(t+1)}  \leftarrow \mathbf{w}^{(t)} - \alpha_{t} \mathbf{dw}^{(t')}_{k}$  where $\alpha_{t} = \alpha \sigma(t-t') n_k / N$.
\ENDFOR
\STATE \underline{\textbf{Middle Layer Aggregator:}}
    \STATE Receive $\big(\mathbf{w}, t''\big)$ from its parent, broadcast to its direct children.
    \STATE Receive $\big(\mathbf{w}_j^{(t')}, t'\big)$ from any of its direct child $j$.
    \STATE Aggregate the collected gradients: $ \mathbf{dw}_k^{(t')} = \sum \alpha \sigma(t''-t') \mathbf{dw}_j^{(t)}$.
    \STATE Send triples $\big(\mathbf{dw}_k^{(t')}, t', n_k\big)$ to its parent.
    
\STATE \underline{\textbf{Bottom Layer Client Device:}}
    \STATE Receive $\big(\mathbf{w}, t''\big)$ from its parent.
    \STATE Define $g_j(\mathbf{w}_j) = \ell_j(\mathbf{w}_j) + \frac{\lambda}{2} || \mathbf{w}_j - \mathbf{w} ||^2 $
    \FOR{local iteration}
        \STATE Randomly sample $(\mathbf{x}, y) \sim \mathcal{D}^i$
        \STATE Local update $\mathbf{w}_j \leftarrow \mathbf{w}_j - \alpha \nabla g_j$
    \ENDFOR
    \STATE Send updated model $\big(\mathbf{w}_j, t''\big)$ to its parent.

\end{algorithmic}
\end{algorithm}

\section{Experiments and Results} \label{sec:res}

\begin{figure*}[ht!]
    \centering
    \begin{subfigure}[b]{0.3\textwidth}
        \centering
        \includegraphics[width=\textwidth]{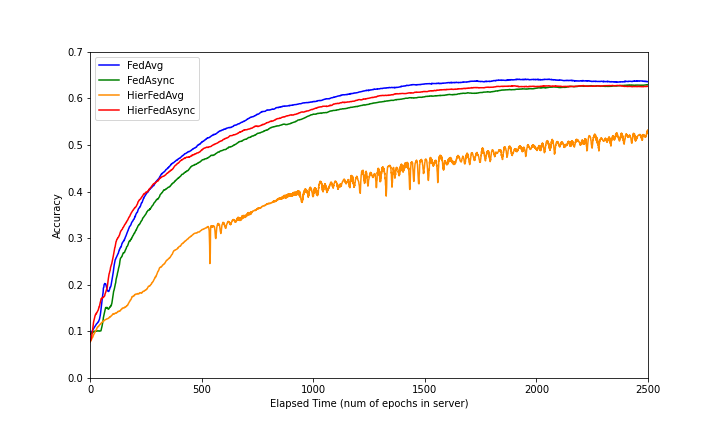}
        \caption{10 client devices, with 2 cluster aggregators for hierarchical learning.}
    \end{subfigure}
    ~
    \begin{subfigure}[b]{0.3\textwidth}
        \centering
        \includegraphics[width=\textwidth]{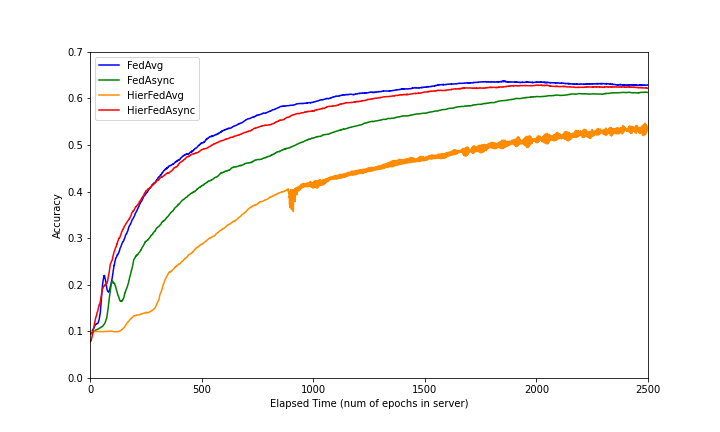}
        \caption{20 client devices, with 4 cluster aggregators for hierarchical learning.}
    \end{subfigure}
    ~
    \begin{subfigure}[b]{0.3\textwidth}
        \centering
        \includegraphics[width=\textwidth]{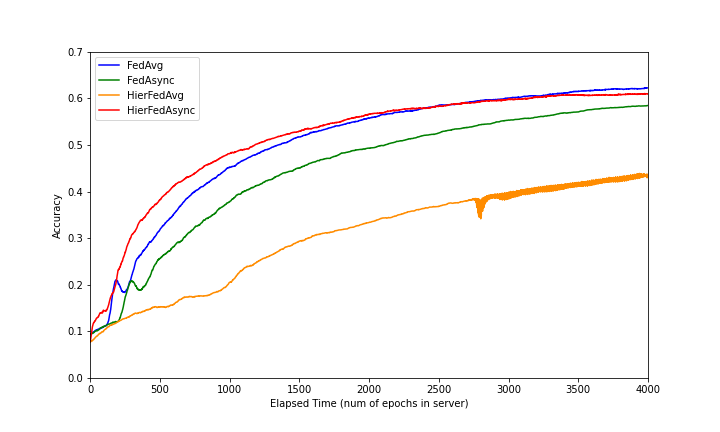}
        \caption{50 client devices, with 5 cluster aggregators for hierarchical learning.}
    \end{subfigure}
    \caption{Comparison of test accuracy w.r.t. training clock in central server.}
    \label{fig:cmp}
\end{figure*}

In our experiments, we consider an asynchronous hierarchical FL system with $N_c$ client devices, $N_a$ cluster aggregators, and a single central server. The non-server machines are grouped into $N_a$ clusters using hierarchical agglomerate cluster based on their IP addresses, while in reality the computational power and network conditions can be integrated as the clustering features as well. For simplicity purpose, the aggregators in each cluster are randomly selected. We conduct our initial experiments on a standard image classification task, the famous CIFAR-10 dataset were used. We set up the model as a convolutional neural network (CNN) with 3 convolutional blocks, which has 5852170 parameters and achieves 90\% test accuracy in centralized training. For our FL system, we randomly partition the CIFAR-10 dataset among the $N_c$ local learning devices, so that each of the 10 class labels are kinds of balanced distributed over all clients. For local training, SGD optimizer are employed with a batch size of 128 and an initial learning rate of 0.001. 

Our models and learning procedures are implemented using PyTorch, and we compare the performance of several different algorithms with our proposed approach. 
Figure \ref{fig:cmp} shows  the comparison of the global models' accuracy evaluated on central servers' validation dataset verses the training time. In \textit{FedAvg} and \textit{FedAsync}, the worker devices directly communicate to the central server without hierarchical structure, while \textit{FedAsync} allows the server and workers to update the models at any time without synchronization. Our hierarchical learning involves the simplest Client-Aggregator-Server 3-layer hierarchical structure, where \textit{HierFedAvg} perform synchronous learning, while \textit{HierFedAsync} follows the learning schema we described in Section \ref{sec:our_method}. 

In each setting, we let the learning last for 2500 learning epochs with a clock in the central server. Specifically, 
each learning epoch is synchronized among all devices. We simulate asynchronous learning system by assuming the fault (e.g. device down, communication loss, straggler effect, etc.) uniformly distributed with probability 0.1 among all non-server devices for each learning epoch. This probability distribution can be further investigated by tuning the parameters and looking at empirical studies. Note that this setting of fault is not introduced into the synchronous learning systems, otherwise a faulty device or network partition may cause the systems to wait forever by their synchronous nature. This in turn demonstrates the advantage of fault tolerance with asynchronous learning.

According to Figure \ref{fig:cmp}, with hierarchical settings, the complexity of learning system is greatly increased, as a result, the \textit{HierFedAvg} algorithm not only converges the slowest, the learning is also not stable. Conversely, it is obvious that our designed \textit{HierFedAsync} algorithm overcomes the issues brought by the hierarchical setting.
Although \textit{FedAvg} seems to perform the best when the number of client devices is small (e.g., when there are 10 clients in the system), we would like to emphasize again that we did not count the fault device nor the stragglers' effect in the synchronous settings for our experiments, while those effects are included in our asynchronous settings. Even so, our \textit{HierFedAsync} algorithm performs close to the best in all cases, and when the system gets larger, the advantages of \textit{HierFedAsync} gets more obvious, not only shows faster convergence especially early on, also leads to higher test accuracy. And we expect further that as the number of devices gets larger, the advantage gets bigger.


Table \ref{tab:num_packets} presents the total numbers of gradients sent or received by each type of devices, from which we can see that the communication burden of the central server would be greatly alleviated in a hierarchical topology, not to mention the potential benefits of more local computation and faster overall convergence. In a large system of network topology, this could also leads to less packet loss, more effective communication and computation.

\begin{table}[]
    \centering
     \caption{Comparison of the numbers of gradients sent/received, with 20 client devices in the system and 2500 training epochs in the central server.}
    \label{tab:num_packets}
    \resizebox{.475\textwidth}{!}{
        \begin{tabular}{l|cccc}
        \hline 
         & \textit{FedAvg} & \textit{FedAsync} & \textit{HierFedAvg} &\textit{HierFedAsync} \\ \hline
        Central Server & 50000 & 44769 & 10000 & 8842 \\ 
        Cluster Aggregators & - & - & 60000 & 52904 \\ 
        Local Clients & 50000 & 45033 & 50000 & 44977 \\ \hline 
        \end{tabular}
    }
\end{table}

\begin{figure}[ht!]
    \centering
    \includegraphics[width=.375\textwidth]{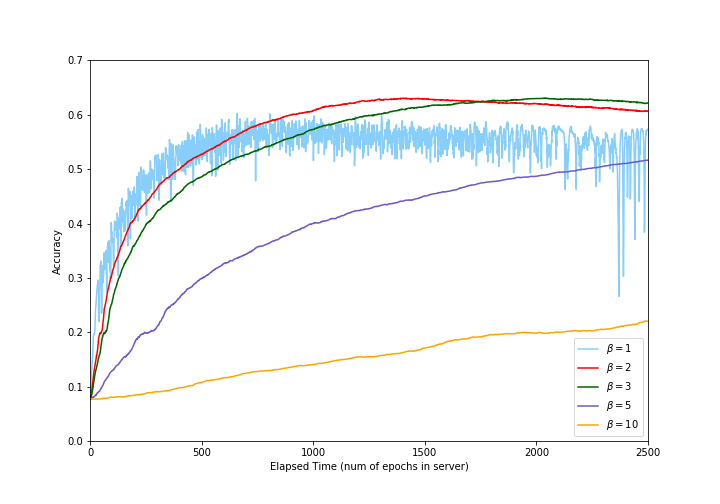}
    \caption{Test accuracy with different $\beta$ values in \textit{HierFedAsync} with 20 client devices and 2500 training epochs in the central server. The polynomial staleness function $\sigma (t'-t) = (t'-t+1)^{-\beta}$ is used on central server as well as cluster aggregators.}
    \label{fig:beta}
\end{figure}

Our \textit{HierFedAsync} algorithm involves several hyperparameters to tune. We conduct comparative analysis with different values of $\beta$ in the polynomial form of staleness function as described in Equation (\ref{eqn:poly_stale}),
and show the effect of staleness on learning convergence in Figure \ref{fig:beta}. We see that when $\beta=1$, the learning curve is closed to that without introducing staleness (i.e. $\beta=0$), but the validation accuracy cannot exceed 55\% as training proceeds, moreover, we notice the learning is unstable as the curve oscillates severely. In general, larger staleness alleviates the instability, at the cost of slower convergence. From Figure \ref{fig:beta}, we can easily see that by using $\beta=2$ or $3$, the performance is significantly improved as the validation accuracy is higher than 60\% after convergence, as the convergence rate is also very acceptable. We also note that  the learning effect is not sensitive with $\beta$ as it is between 2 and 3, indicates that $\beta$ is quite easy to tune. 
Similar comparative analysis is conducted for the regularization coefficient $\lambda$ in Equation (\ref{eqn:local}) on local clients. Although results in Figure 
shows little effect for change of $\lambda$,  we would like to note that our current experimental setting does not emphasize on simulation of the stragglers.
We expect that the regularization on local clients plays a more important role during training in an asynchronous system as the straggler effect gets more common.

\section{Conclusions and Future Work} \label{sec:conc}

In this paper, we propose asynchronous hierarchical federated learning. 
As federated networks are composed of a massive number of devices, communication is a critical challenge in FL. We tackle this problem by exploring different architectural patterns for the design of FL systems. The tradeoff of central and fully decentralized learning on the complexity of the system as well as the learning effectiveness, computational and communication cost is obvious. We deploy a FL system with a central server, but with hierarchical topology. In this paper, we combine asynchronous FL and hierarchical FL into our \textit{FedAH} algorithm. In addition, we blur the concept of network topological edges to form clusters as well as the hierarchical structures. We aim at reducing the communication load between devices and the server in FL system, also improve flexibility and scalability. Our initial experiments demonstrated that combining asynchronous FL and hierarchical FL not only leads to faster convergence, tolerates heterogeneity of the system such as the faulty devices, straggler effect, etc., also significantly alleviates the communication burden on the central server. However, the asynchronous and hierarchical nature greatly increases the complexity of the system, especially on the communication topology, which could lead to unstable learning. 
We explored the literature and recent research advances, combine them into our proposed method, \textit{FedAH}, which inherits the merits of both asynchronous and hierarchical FL, meanwhile significantly mitigates the instability with the utilization of staleness function and cluster weighting on the central server and edge devices, as well as adding regularization for local updates on client devices. We implemented the system and evaluated on CIFAR-10 image classification task, the results verify the effectiveness of our design and meet our expectation. 
There are several interesting directions to pursue for the future of this paper. First off, as we mentioned in the paper, our weighting mechanism for aggregation favors learning for faster computed and communicated devices, which works fine if the data are i.i.d. over the clients. We need to design a more sophisticated weighting mechanism for asynchronous FL if non-i.i.d. data are involved. 
The second interesting direction in which to take this paper would be modification of the simple $L^2$-regularization on local clients' learning.
We also need to finish the derivation and proof of the theoretical analysis. 
In addition, more experiments need to be conducted on different datasets and tasks. 
Furthermore, we need to bring more sophisticated experimental settings for the simulation, for instance, the straggler effect should be emphasized.






{\small
\bibliographystyle{abbrvnat}
\bibliography{FlRef}
}

\end{document}